# Evolutionary Algorithm for Drug Discovery – Interim Design Report

Author: Mark Shackelford                                       14[th] March 2014

Post:   Advanced Computational Technologies
        9 Belsize Road, West Worthing, West Sussex, BN11 4RH, UK
Email:  mark_shackelford@hotmail.com

**Introduction**
Evolutionary Algorithm for Drug Discovery [EAfDD] is a MS Windows based software program (written in C#.NET) which aims to provide an explorative capability over the Search Space of potential drug molecules.

In essence the program explores the search space by generating random molecules, determining their fitness (based on a variety of criteria) and then breeding a new generation from the fittest individuals. The search space (in theory any combination of any elements in any order) is constrained by the use of a subset of elements (e.g. organic) and a list of fragments (molecular parts that are known to be useful in drug development).

The resultant molecules from each generation are stored in a searchable database, so that the user can browse through previous generations looking for interesting molecules.

Much of the intent of the program is not necessarily to find an optimal solution, or even a molecule that can actually be created by a chemist – but to allow the user to trawl through areas of the search space that might be outside their experience.

The program also provides an Interactive capability, whereby the user can inspect the output from any generation and adjust the system calculated fitness, thereby allowing the user to alter the selection process and drive the algorithm in a new direction.

There have been a number of different approaches in the past ([1] – [35]), which have used fixed length genetic algorithms, or graph based genetic programs etc. These have produced variable success but have provided many useful ideas that been incorporated into EAfDD.

**The Algorithm**
EAfDD uses an evolutionary algorithm with variable length strings to represent candidate molecules. The initial population of molecules is created using a newly designed molecule creation process, which builds a C# object-oriented linked-list structure of molecular fragments and atoms. The fragments and atoms are selected from a user-specifiable list of required items – which can be varied depending on the drug target environment.

Standard chemistry rules (valence etc) are used to create the initial molecules, and a variety of other rules can also be specified and used to determine the validity of the molecules (e.g. min/max number of atoms, max molecular weight etc.).

Once the program has created the molecules they are converted to "expanded" SMILES strings for processing within the evolutionary algorithm – which has a variety of functions to search and select items within the strings for cross-over and mutation.

NOTE: "Expanded" SMILES is an extended format of the basic SMILES strings, which contains extra information to allow the cross-over and mutation operators to easily find the required items to process. For details see Expanded SMILES section below.

The resultant population runs through a number of iterations using a genetic algorithm cross-over and mutation operators, which have been tailored to meet the chemical rules as specified with the program.

The Fitness of molecules can be derived from a variety of mechanisms, such as using algorithms to determine solubility, permeability, toxicity etc. At present the algorithm (in proof of concept mode) uses a Target Molecule and a Similarity Fitness function, such that it has to evolve its initial random population towards a target molecule. Early tests show that the algorithm can reach a 96% fit with the target molecule in 100 iterations or so.

**The Program**

The program provides a range of functions to allow the process to be tailored to a wide range of drug discovery projects.

- Parameters – this function allows the user to control all aspects of the process, from the number of iterations, the various GA parameters (population size etc), the generation process (such as the % likelihood of choosing a new atom or fragment when creating a molecule) and the cross-over and mutation processes (such as mutation rate).

- Elements – the system allows the user to choose a subset of elements to be used in the process. This is likely to be the Organic subset (H, B, C, N, O, F, P, S, Cl, Br, I), but this can be extended or reduced as required. A Periodic Table graphic is used to select the required atoms.

- Fragments – the user can create a list of molecular fragments that are to be used in the random molecule generation. These are specified as SMILES strings and will be used in the initial population as well as during the Mutation process.

- Leads – one or more Lead Molecules can be specified to seed the initial population. Use of this function will tend to target specific areas of the search space, and can be used to explore potential drug molecules similar to known ones, rather than the more general exploration of the entire search space.

- Fitness – a variety of operators can be specified for use in defining the fitness of molecules which drives the selection process for each generation. At present this uses a Similarity function (how close to a target molecule) plus constraints on the size of molecules (number of atoms and their molecular weight).

- History – all generated molecules for every generation are saved in a SQL database, and can be accessed through a search and display function. This allows the user to not only be given the final "optimised" molecule, but browse through all of the generated molecules to see if there are other interesting molecules that have potential even if they have not met the current fitness criteria.

- Interaction – as the program iterates through the generations, the user can break into the process, review each molecule (the system draws a 2D representation of each molecule on the screen and provides a variety of data on the molecule (weight etc.). The user can then alter the fitness score of any or all of the molecules, and then let the process continue. As the fitness score drives the selection process, the iteractive user is able to drive the evolution towards their own required direction – which may not be able to be defined with the fitness functions provided.

**Selection Operators**

The system provides 6 different Parent Selection methods for the genetic algorithm Cross-Over process, the user deciding which one is the most appropriate for their particular requirements:

- Standard (Roulette) - each Chromosome is chosen based on its relative Fitness. The higher the fitness the more likely it will be chosen as a parent.

- Tournament - a number of Chromosomes are chosen at random, and the two highest Fitness chromosomes are used as Parents

- Random - two Parents are chosen at random with no bearing on fitness

- Attractive - a number of Chromosomes are chosen at random and the two with the closest Similarity are chosen

- Difference - a number of Chromosomes are chosen at random and the two with the least Similarity are chosen

- Sequential - each Chromosome is chosen in order as a Parent, with a second one chosen at random.

The last three Operators are experimental, and not previously described in Evolutionary Algorithm literature - but would seem to have the potential to model biological behaviour - attraction of like, attraction of opposites - or free-for-all.

**Mutation Operators**

The system provides 10 mutation operators, which are chosen at random for each iteration and each member of the population. If one mutation process fails, the next random one is attempted until a successful mutation occurs (or all operators have been used).

A Mutation Rate (%) parameter is used to determine whether any specific Molecule is subjected to Mutation.

- Insert Atom - a bond (-, =) is chosen at random and a new Atom/Fragment with the same Valence is inserted.

- Replace H - a single H atom is selected and replaced with a branching Atom/Fragment

- Remove Atom - an Atom is replaced by one or more [H] to match Valence

- Remove Bond and Atom - an Atom with equal bonds is replaced by a single bond

- Change Atom to Fragment - an Atom at the end of a branch is replaced with a Fragment with the same Valence

- Switch Atom - change one Atom for another with same or greater Valence (plus H)

- Increase Bond - a single bond with available H at both ends is replaced with a double bond, and an H removed from each adjoining Atom

- Decrease Bond - a double (or triple) bond is replaced with a smaller bond, with H added to each adjoining Atom

- Cut Ring - two matching Ring markers {n} are removed and replaced with [H]

- Add Ring - two H atoms at the ends of branches are replaced with {n}

**User Interaction**

The system provides a variety of Interactive options for the user. These allow any individual from each generation to be rescored by the user. Scoring is very basic – allowing a set of values in the range {0, 1} to be defined. For example these might be "Excellent = 1, Good = 0.75, Average = 0.5, Poor = 0.25, Dreadful = 0.00". The range and values can be defined in the interface.

The user can choose the interval at which the system allows interaction – i.e. Interaction every <n> generations, which can range from 1 (every generation) to 100 (every 100[th] generation).

The system provides a number of options for which individuals are displayed for interaction, ranging from ALL, to selected individuals based on a number of optional criteria:

- Top <n> - only the top individual are displayed, based on fitness score. This can have a deleterious effect on diversity,

- Banding - using a normal sized population, split into (N) bands (where N is a suitable size to present) across the fitness spectrum. Display one item (at random) from each Band, and then use the input User's score to scale the original band scoring

- Partial Sequential - use a normal sized population but select only some by choosing every nth item in fitness order. User scores these ones, system scores the remaining.

- Partial Random - use a normal sized population, and select (N) at random for the user to score. Rest scored by the system.

The last three attempt to maintain diversity, whilst allowing the user to have some effect (but not completely). It was shown (*Shackelford & Corne, 2005*) that the User often makes mistakes, or loses concentration - so it is useful to have the machine scoring in parallel.

**Expanded SMILES**

SMILES is a standard (developed by Daylight Systems) for specifying a molecular formula as a simple linear string. This format (in itself) does not contain enough information for the algorithm's Crossover and Mutation operators, so the system uses an "Expanded SMILES" notation - which can easily be converted to standard SMILES for use with the Chemaxon function library.

The rules for generating Expanded SMILES are as follows:

All atoms (apart from Hydrogen) to be specified within square brackets: [C]

All bonds to be entered explicitly: [C]-[N], [C]=[O], [C]#[N] where "-" is single, "=" is double and "#" is triple

Hydrogen atoms to be explicitly added as required inside their bonded atom [XH], followed by number unless 1: [CH2], [OH]

All branches ending in zero valence to be entered within round brackets (): [C](=[O])-[C](-[F])(=[O])

Ring indication numbers must be shown with curly brackets: [C](-[O]-{1})(=[CH]-[C](=[N]-{1})-[OH]

Ring indicators must also include the bond symbol (-, =, #)